\documentclass{article}
\usepackage[utf8]{inputenc}
\pdfoutput=1
\usepackage[sc,osf]{mathpazo}
\usepackage[margin=1.1in,letterpaper]{geometry}
\usepackage[colorlinks=true,citecolor=blue,breaklinks]{hyperref}
\usepackage[hyphenbreaks]{breakurl} 
\usepackage[lined,boxed]{algorithm2e}
\usepackage{cite}

\makeatletter
\newlength\aftertitskip     \newlength\beforetitskip
\newlength\interauthorskip  \newlength\aftermaketitskip

\def\maketitle{\par
 \begingroup
   \def\thefootnote{\fnsymbol{footnote}}
   \def\@makefnmark{\hbox to 4pt{$^{\@thefnmark}$\hss}}
   \@maketitle \@thanks
 \endgroup
\setcounter{footnote}{0}
 \let\maketitle\relax \let\@maketitle\relax
 \gdef\@thanks{}\gdef\@author{}\gdef\@title{}\let\thanks\relax}

\def\@startauthor{\noindent \normalsize\bf}
\def\@endauthor{}
\def\@starteditor{\noindent \small {\bf Editor:~}}
\def\@endeditor{\normalsize}
\def\@maketitle{\vbox{\hsize\textwidth
 \linewidth\hsize \vskip \beforetitskip
 {\begin{center} \LARGE\@title \par \end{center}} \vskip \aftertitskip
 {\def\and{\unskip\enspace{\rm and}\enspace}%
  \def\addr{\small\it}%
  \def\email{\hfill\small\tt}%
  \def\name{\normalsize\bf}%
  \def\AND{\@endauthor\rm\hss \vskip \interauthorskip \@startauthor}
  \@startauthor \@author \@endauthor}
}}

\makeatother

\pdfoutput=1                    

\usepackage{floatrow} 
\newfloatcommand{capbtabbox}{table}[][\FBwidth]

\usepackage{graphicx}
\usepackage{amsmath,amsthm,amssymb,bm} 
\usepackage{amsfonts}
\usepackage{xspace}
\usepackage{array}
\usepackage{enumerate}

\numberwithin{equation}{section}

\newcommand{\Sp}{\mathbb{S}}
\newcommand{\Spd}{\Sp^{p-1}}
\newcommand{\reals}{\mathbb{R}}
\newcommand{\rbar}{\bar{r}}
\newcommand{\m}{\mu}
\newcommand{\vmfd}{p_{\text{vmf}}}
\newcommand{\watson}{p_{\text{wat}}}

\newcommand{\norm}[1]{\left\Vert #1\right\Vert}
\newcommand{\kummer}{M}

\newcommand{\risingF}[2]{{{#1}}^{\overline{{#2}}}}
\newcommand{\half}{\tfrac{1}{2}}
\newcommand{\khat}{\hat{\kappa}}
\newcommand{\ip}[2]{\left\langle {#1},\, {#2}\right\rangle}
\newcommand{\set}[1]{\{{#1}\}}

\newcommand{\ratio}{\frac{M'(a, c; \kappa)}{M(a, c; \kappa)}}
\newcommand{\mh}{\hat{\m}}
\newcommand{\proj}{\mathbb{P}^{p-1}}

\newcommand{\nlsum}{\sum\nolimits}
\numberwithin{equation}{section}

\DeclareMathOperator{\trace}{Tr}
\DeclareMathOperator*{\argmax}{argmax}
\newtheorem{theorem}{Theorem}[section]

\begin{document}
\title{Directional Statistics in Machine Learning: a Brief Review\thanks{Invited chapter submitted to: ``Modern Statistical Methods for Directional Data'' (eds. C. Ley and T. Verdebout)}}
\author{\name Suvrit Sra \email{suvrit@mit.edu}\\
  \addr{Massachusetts Institute of Technology, Cambridge, MA 02139} 
}

\maketitle

\begin{abstract}
  The modern data analyst must cope with data encoded in various forms, vectors, matrices, strings, graphs, or more. Consequently, statistical and machine learning models tailored to different data encodings are important. We focus on data encoded as normalized vectors, so that their ``direction'' is more important than their magnitude. Specifically, we consider high-dimensional vectors that lie either on the surface of the unit hypersphere or on the real projective plane. For such data, we briefly review common mathematical models prevalent in machine learning, while also outlining some technical aspects, software, applications, and open mathematical challenges. \end{abstract}

\section{Introduction}
Data are often represented as vectors in a Euclidean space $\reals^p$, but frequently, data possess more structure and treating them as Euclidean vectors may be inappropriate. A simple example of this instance is when data are normalized to have unit norm, and thereby put on the surface of the \emph{unit hypersphere} $\Spd$. Such data are better viewed as objects on a manifold, and when building mathematical models for such data it is often advantageous to exploit the geometry of the manifold (here $\Spd$).

For example, in classical information retrieval it has been convincingly demonstrated that cosine similarity is a more effective measure of similarity for analyzing and clustering text documents than just Euclidean distances. There is substantial empirical evidence that normalizing the data vectors helps to remove the biases induced by the length of a document and provide superior results~\cite{salton83,salton89}. On a related note, the spherical k-means (\texttt{spkmeans}) algorithm~\cite{spkmeans} that runs k-means with cosine similarity for clustering unit norm vectors, has been found to work well for text clustering and a variety of other data. Another widely used similarity measure  is \emph{Pearson correlation}: given $x, y \in \reals^d$ this defined as $\rho(x,y) := \frac{\sum_i(x_i-\bar{x})(y_i-\bar{y})}{\sqrt{\sum_i(x_i-\bar{x})^2}\times \sqrt{\sum_i (y_i-\bar{y})^2}},$ where $\bar{x} = \frac1d\sum_i x_i$ and $\bar{y}=\frac1d\sum_iy_i$. Mapping $x \mapsto \tilde{x}$ with $\tilde{x}_i = \frac{x_i-\bar{x}}{\sqrt{\sum_i(x_i-\bar{x})^2}}$ (similarly define $\tilde{y}$), we obtain the inner-product $\rho(x,y) = \ip{\tilde{x}}{\tilde{y}}$. Moreover, $\norm{\tilde{x}}=\norm{\tilde{y}}=1$. Thus, the Pearson correlation is exactly the cosine similarity between $\tilde{x}$ and $\tilde{y}$.  More broadly, domains where similarity measures such as cosine, Jaccard or Dice~\cite{rasmussen92} are more effective than measures derived from Mahalanobis type distances, possess intrinsic ``directional'' characteristics, and are hence better modeled as directional data~\cite{mardia75b}.

This chapter recaps basic statistical models for \emph{directional data}, which herein refers to unit norm vectors for which ``direction'' is more important than ``magnitude.'' In particular, we recall some basic distributions on the unit hypersphere, and then discuss two of the most commonly used ones: the von Mises-Fisher and Watson distributions. For these distributions, we describe maximum likelihood estimation as well as mixture modeling via the Expectation Maximization (EM) algorithm. In addition, we include a brief pointer to recent literature on applications of directional statistics within machine learning and related areas.

We warn the advanced reader that no new theory is developed in this chapter, and our aim herein is to merely provide an easy introduction. The material of this chapter is based on the author's thesis~\cite{sra.thesis}, and the three papers~\cite{sra.vmf,sra.mow,sra09}, and the reader is referred to these works for a more detailed development and additional experiments.


\section{Basic Directional Distributions}
\subsection{Uniform distribution}
The probability element of the uniform distribution on $\Spd$ equals $c_pd\Spd$. The  normalization constant $c_p$ ensures that $\int_{\Spd} c_p d\Spd = 1$, from which it follows that 
\begin{equation*}
  c_p = \Gamma(p/2)/2\pi^{p/2},
\end{equation*}
where $\Gamma(s) := \int_0^\infty t^{s-1}e^{-t}dt$ is the well-known Gamma function.

\subsection{The von Mises-Fisher distribution}
The vMF distribution is one of the simplest distributions for directional data and it has properties analogous to those of the multivariate Gaussian on $\reals^p$. For instance, the maximum entropy density on $\Spd$ subject to the constraint that $E[x]$ is fixed, is a vMF density (see e.g., \cite[pp. 172--174]{rao73} and \cite{mardia75}).

A unit norm vector $x$ has the von Mises-Fisher (vMF) distribution if its density is 
\begin{equation*}
  \vmfd(x; \mu, \kappa) := c_p(\kappa) e^{\kappa \mu^Tx},
\end{equation*}
where $\|\mu\| = 1$ and $\kappa \ge 0$. Integrating using polar coordinates, it can be shown~\cite[App.~B.4.2]{sra.thesis} that the normalizing constant is given by
\begin{equation*}
  c_p(\kappa) = \frac{\kappa^{p/2-1}}{(2\pi)^{p/2} I_{p/2 - 1}(\kappa)},
\end{equation*}
where $I_s(\kappa)$ denotes the modified Bessel function of the first kind~\cite{abst74}.\footnote{Note that sometimes in directional statistics literature,  the integration measure is normalized by the uniform measure, so that
instead of $c_p(\kappa)$, one uses $c_p(\kappa) 2\pi^{p/2} / \Gamma(p/2)$.}

The vMF density $\vmfd = c_p(\kappa)e^{\kappa \mu^T x}$ is parameterized by the mean direction $\mu$, and the \emph{concentration} parameter $\kappa$, so-called because it characterizes how strongly the unit vectors drawn
according to $\vmfd$ are concentrated about the mean direction $\mu$.  Larger values of $\kappa$ imply stronger concentration about the mean direction.  In particular when $\kappa=0$, $\vmfd$ reduces to the uniform density on $\Spd$, and as $\kappa\to \infty$, $\vmfd$ tends to a point density.

\subsection{Watson distribution}
The uniform and the vMF distributions are defined over \emph{directions}.  However, sometimes the observations are \emph{axes} of direction, i.e., the vectors $\pm x \in \Spd$ are equivalent. This constraint is also denoted by $x \in \proj$, where $\proj$ is the projective hyperplane of dimension $p{-}1$. The multivariate Watson distribution~\cite{maju00} models such data; it is parametrized by a
\emph{mean-direction} $\mu \in \proj$, and a \emph{concentration} parameter
$\kappa \in \reals$, with probability density
\begin{equation}
  \label{eq:60}
  \watson(x; \mu, \kappa) := d_p(\kappa) e^{\kappa (\mu^T x)^2},\qquad  x \in \proj.
\end{equation}
The normalization constant $d_p(\kappa)$ is given by
\begin{equation}
  \label{eq:one}
  d_p(\kappa) = \frac{\Gamma(p/2)}{2\pi^{p/2}\kummer(\half, \tfrac{p}{2}, \kappa)},
\end{equation}
where $\kummer$ is the confluent hypergeometric function \cite[formula~6.1(1)]{htf}
\begin{equation}
  \label{eq:two}
  \kummer(a, c, \kappa) = \sum_{j \geq 0} \frac{\risingF{a}{j}}{\risingF{c}{j}} \frac{\kappa^j}{j!},\qquad a, c, \kappa \in \reals,
\end{equation}
and $\risingF{a}{0}=1$, $\risingF{a}{j}=a(a+1)\cdots(a+j-1)$,
$j\geq{1}$, denotes the \emph{rising-factorial}.

Observe that for $\kappa > 0$, the density concentrates around $\mu$ as $\kappa$ increases, whereas for $\kappa < 0$, it concentrates around the great circle orthogonal to $\mu$. 

\subsection{Other distributions}
We briefly summarize a few other interesting directional distributions, and refer the reader to~\cite{maju00} for a more thorough development.

\paragraph{Bingham distribution.}
Some axial data do not exhibit the rotational symmetry of Watson distributions. Such data could be potentially modeled using Bingham distributions, where the density at a point $x$ is $B_p(x; K) := c_p(K)e^{x^TKx}$, where the normalizing constant $e_p$ can be shown to be $c_p(K) = \frac{\Gamma(p/2)}{2\pi^{p/2}\kummer(\half, \tfrac{p}{2}, K)}$, where $\kummer(\cdot, \cdot, K)$ denotes the confluent hypergeometric function of matrix argument~\cite{muirhead82}. 

Note that since $ x^T(K + \delta I_p) x =  x^TKx + \delta$, the Bingham density is identifiable only up to a constant diagonal shift. Thus, one can assume $\trace(K) =
0$, or that the smallest eigenvalue of $K$ is zero~\cite{maju00}. Intuitively, one can see that the eigenvalues of $K$ determine the axes around which the data clusters, e.g., greatest clustering will be around the axis corresponding to the leading eigenvector of $K$.


\paragraph{Bingham-Mardia distribution.}
Certain problems require rotationally symmetric distributions that have a `modal ridge'
rather than just a mode at a single point. To model data with such characteristics \cite{maju00} suggest a density of the form
\begin{equation}
  \label{eq:33}
  p(x; \mu, \kappa, \nu) = c_p(\kappa) e^{\kappa (\mu^T x-\nu)^2},
\end{equation}
where as usual $c_p(\kappa)$ denotes the normalization constant.

\paragraph{Fisher-Watson distributions}
This distribution is a simpler version of the more general Fisher-Bingham distribution~\cite{maju00}. The density is
\begin{equation}
  \label{eq:34}
  p(x; \mu, \mu_0, \kappa, \kappa_0) = c_p(\kappa_0, \kappa, \mu_0^T\mu) 
  e^{\kappa_0\mu_0^T x + \kappa (\mu^T x)^2}.
\end{equation}

\paragraph{Fisher-Bingham.}
This is a more general directional distribution; its density is
\begin{equation}
  \label{eq:32}
  p(x; \mu, \kappa, A) = c_p(\kappa, A) e^{\kappa\mu^T x +  x^TAx}.
\end{equation}
There does not seem to exist an easy integral representation of the normalizing constant, and in an actual application one needs to resort to some sort of approximation for it (such as a saddle-point approximation). Kent distributions arise by putting an additional constraint $A\mu = 0$ in \eqref{eq:32}.

\section{Related work and applications}
\label{sec:rel}
The classical references on directional statistics are~\cite{mardia75,mardia75b,watson1966statistics}; a more recent, updated reference is~\cite{maju00}. Additionally, for readers interested in statistics on manifolds, a good starting point is~\cite{chikuse2012statistics}. To our knowledge, the first work focusing on high-dimensional application of directional statistics was~\cite{sra.vmf}, where the key application was clustering of text and gene expression data using mixtures of vMFs. There exist a vast number of other applications and settings where hyperspherical or manifold data arise. Summarizing all of these is clearly beyond the scope of this chapter. We mention below a smattering of some works that are directly related to this chapter.

We note a work on feature extraction based on correlation in~\cite{fu2008correlation}. Classical data mining applications such as topic modeling for normalized data are studied in~\cite{banerjee2007topic,reisinger2010spherical}. A semi-parametric setting using Dirichlet process mixtures for spherical data is~\cite{straub2015dirichlet}. Several directional data clustering settings include: depth images using Watson mixtures~\cite{hasnat2014unsupervised}; a k-means++~\cite{kmpp} style procedure for mixture of vMFs~\cite{mash2015k};  clustering on orthogonal manifolds~\cite{cetingul2009intrinsic}; mixtures of Gaussian and vMFs~\cite{kasarapu2015minimum}. Directional data has also been used in several biomedical (imaging) applications, for example \cite{mcgraw2006mises}, fMRI~\cite{lashkari2010discovering}, white matter supervoxel segmentation~\cite{cabeen2014white}, and brain imaging~\cite{ryali2013parcellation}. In signal processing there are applications to spatial fading using vMF mixtures~\cite{mammasis2009spatial} and speaker modeling~\cite{tang2009generative}. 
Finally, beyond vMF and Watson, it is worthwhile to consider the Angular Gaussian distribution~\cite{tyler1987statistical}, which has been applied to model natural images for instance in~\cite{hosseini.thesis}. 

\section{Modeling directional data: maximum-likelihood estimation}
\label{sec:params}
In this section we briefly recap data models involving vMF and Watson distributions. In particular, we describe maximum-likelihood estimation for both distributions. As is well-known by now, for these distributions estimating the mean $\mu$ is simpler than estimating the concentration parameter $\kappa$.

\subsection{Maximum-Likelihood estimation for vMF}
\label{sec:kappa}
Let ${\cal X} = \{x_1,\ldots, x_n\}$ be a set of points drawn from $\vmfd(x;\mu,\kappa)$. We wish to estimate $\mu$ and $\kappa$ by solving the m.l.e.\ optimization problem
\begin{equation}
  \label{eq:4}
  \max \ell({\cal X}; \mu, \kappa) := \log c_p(\kappa) + \nlsum_{i=1}^n \kappa \mu^T x_i,\quad\text{s.t.}\ \norm{\mu}=1,\ \kappa \ge 0.
\end{equation}
Writing $\frac{\|\sum_i x_i\|}{n} = \rbar$, a brief calculation shows that the optimal solution satisfies
\begin{equation}
  \label{eq:7}
    \mu = \tfrac{1}{n\rbar}\nlsum_{i=1}x_i,\quad \kappa = A_p^{-1}(\rbar),
\end{equation}
where the nonlinear map $A_p$ is defined as
\begin{equation}
  \label{eq:8}
  A_p(\kappa) = \frac{-c_p'(\kappa)}{c_p(\kappa)} = \frac{I_{p/2}(\kappa)}{I_{p/2-1}(\kappa)} = \rbar.
\end{equation}
The challenge is to solve~\eqref{eq:8} for $\kappa$. For small values of $p$ (e.g., $p=2,3$) the simple estimates provided in~\cite{maju00} suffice. But for machine learning problems, where $p$ is typically very large, these estimates do not suffice. In~\cite{sra.vmf}, the authors provided efficient numerical estimates
for $\kappa$ that were obtained by truncating the continued fraction representation
of $A_p(\kappa)$ and solving the resulting equation. These estimates were then corrected to yield the approximation
\begin{equation}
  \label{eq:1}
  \hat{\kappa} = \frac{\rbar(p - \rbar^2)}{1-\rbar^2},
\end{equation}
which turns out to be remarkably accurate in practice.

Subsequently,~\cite{tanabe} showed simple bounds for $\kappa$ by exploiting
inequalities about the Bessel ratio $A_p(\kappa)$---this ratio possesses several
nice properties, and is very amenable to analytic treatment~\cite{amos74}. The
work of~\cite{tanabe} lends theoretical support to the empirically determined
approximation~\eqref{eq:1}, by essentially showing this approximation lies in the ``correct'' range. Tanabe et al.~\cite{tanabe} also presented a
fixed-point iteration based algorithm to compute an approximate solution $\kappa$.

The \emph{critical} difference between this approximation and the next two is
that it does not involve any Bessel functions (or their ratio). That is, not a
single evaluation of $A_p(\kappa)$ is needed---an advantage that can be significant
in high-dimensions where it can be computationally expensive to compute
$A_p(\kappa)$. Naturally, one can try to compute $\log I_s(\kappa)$ ($s=p/2$) to avoid
overflows (or underflows as the case may be), though doing so introduces yet
another approximation. Therefore, when running time and simplicity are of the
essence, approximation~\eqref{eq:1} is preferable.

Approximation~\eqref{eq:1} can be made more exact by performing a few iterations of Newton's method. To save runtime, \cite{sra09} recommends only two-iterations of Newton's method, which amounts to computing $\kappa_0$ using~\eqref{eq:1}, followed by
\begin{equation}
  \label{eq:2}
  \begin{split}
    \kappa_{s+1} = &\ \kappa_{s} - \frac{A_p(\kappa_{s}) - \bar{R}}{1 - A_p(\kappa_{s})^2 - \frac{(p-1)}{\kappa_s}A_p(\kappa_s)},\quad s=0,1.
  \end{split}
\end{equation}
Approximation~\eqref{eq:2} was shown in~\cite{sra09} to be competitive in running time with the method of~\cite{tanabe}, and was seen to be overall more accurate. Approximating $\kappa$ remains a topic of research interest, as can be seen from the recent works~\cite{hornik2014maximum,christie2014efficient}.

\subsection{Maximum-Likelihood estimation for Watson}
\label{sec:mle}
Let ${\cal X} = \{x_1,\ldots, x_n\}$ be i.i.d.\ samples drawn from $\watson( x;\m,\kappa)$. We wish to estimate $\mu$ and $\kappa$ by maximizing the log-likelihood
\begin{equation}
  \label{eq:3}
  \ell({\cal X}; \mu,\kappa) = n\bigl(\kappa\m^\top{S}\m -
  \ln\kummer(1/2, p/2, \kappa) + \gamma\bigr),
\end{equation}
subject to $\mu^T\mu=1$, where $S = \frac1n\sum_{i=1}^n  x_i x_i^\top$ is the sample \emph{scatter
  matrix}, and $\gamma$ is a constant. Considering the first-order optimality conditions of~\eqref{eq:3} leads to the following parameter estimates~\cite[Sec.~10.3.2]{maju00}
\begin{equation}
  \label{four}
  \hat{\m} = \pm s_1\quad\text{if}\quad \khat > 0,\qquad \mh = \pm s_p\quad\text{if}\quad \khat < 0,
\end{equation}
where $s_1,s_2,\ldots,s_p$ are (normalized) eigenvectors of the scatter matrix $S$ corresponding to the eigenvalues $\lambda_1 \ge \lambda_2 \ge \cdots \ge \lambda_p$. 

To estimate the concentration parameter $\khat$ we must solve:\footnote{We need $\lambda_1 > \lambda_2$ to ensure a unique m.l.e.\ for positive $\kappa$, and $\lambda_{p-1} > \lambda_p$, for negative $\kappa$}
\begin{equation}
  \label{eq:five}
  g(\half,\tfrac{p}{2}; \khat) := \frac{\frac{\partial}{\partial\kappa}\kummer(\half, \tfrac{p}{2}, \khat)}
  {\kummer(\half, \tfrac{p}{2}, \khat)}\ \ =\ \ \mh^\top{S}\mh\ :=\
  r\qquad(0\le r\le 1),
\end{equation}
Notice that~\eqref{four} and~\eqref{eq:five} are coupled---so we simply solve both $g(1/2,p/2;\khat) = \lambda_1$ and $g(1/2, p/2;\khat) =\lambda_p$, and pick the solution that yields a higher log-likelihood. 

The hard part is to solve~\eqref{eq:five}. One could use a root-finding method (e.g.~Newton-Raphson), but similar to the vMF case, an out-of-the-box root-finding approach can be unduly slow or numerically hard as data dimensionality increases. The authors of~\cite{sra.mow} consider the following more general equation:
\begin{equation}
  \label{eq:61}
  \begin{split}
    \quad g(a, c; \kappa) := &\ratio = r\\
    c > a > 0,&\quad 0 \leq r \leq 1,
  \end{split}
\end{equation}
and derive for it high-quality closed form numerical approximations. These approximations improve upon two previous approaches, that of~\cite{bijral} and~\cite{sra.thesis}.  Bijral et al.~\cite{bijral} followed the continued-fraction approach of~\cite{sra.vmf} to obtain the heuristic approximation
\begin{equation}
  \label{eq:bijral}
  BBG(r) := \frac{cr-a}{r(1-r)} + \frac{r}{2c(1-r)}.
\end{equation}
Other heuristic approximations were presented by the author in~\cite{sra.thesis}.

The following theorem of~\cite{sra.mow} provides rigorously justified approximations, most of which are typically more accurate than previous heuristics.
\begin{theorem}[\cite{sra.mow}]
  \label{thm:bounds}
  Let the solution to $g(a, c; \kappa)=r$ be denoted by $\kappa(r)$. Consider the following three bounds:
  \begin{align}
    \label{eq:lower}
    &(\text{lower bound})\qquad &L(r)&=\frac{rc-a}{r(1-r)}\left(1+\frac{1-r}{c-a}\right),\\
    \label{eq:mid}
    &(\text{bound})\qquad &B(r)&=\frac{rc-a}{2r(1-r)}\left(1+\sqrt{1+\frac{4(c+1)r(1-r)}{a(c-a)}}\right),\\
    \label{eq:upper}
    &(\text{upper bound})\qquad &U(r)&=\frac{rc-a}{r(1-r)}\left(1+\frac{r}{a}\right).
  \end{align}
  Let $c > a > 0$, and $\kappa(r)$ be the solution~\eqref{eq:61}. Then, we have
  \begin{enumerate}
  \item for $a/c<r<1$,
    \begin{equation}\label{eq:newpos}
      L(r) < \kappa(r) < B(r) < U(r),
    \end{equation}
  \item for $0<r<a/c$,
    \begin{equation}\label{eq:newneg}
      L(r) < B(r) < \kappa(r) < U(r).
    \end{equation}
  \item and if $r=a/c$, then $\kappa(r) = L(a/c)=B(a/c)=U(a/c)=0$.
  \end{enumerate}
  All three bounds ($L$, $B$, and $U$) are also
  asymptotically precise at $r=0$ and $r=1$.
\end{theorem}

\section{Mixture models}
\label{sec:mix}
Many times a single vMF or Watson distribution is insufficient to model data. In these cases, a richer model (e.g., for clustering, or as a generative model, etc.) such as a mixture model may be more useful. We summarize in this section mixtures of vMF (movMF) and mixtures of Watson (moW) distributions. The former was originally applied to high-dimensional text and gene expression data in~\cite{sra.vmf}, and since then it has been used in a large number of applications (see also Section~\ref{sec:rel}). The latter has been applied to genetic data~\cite{sra.mow}, as well as in other data mining applications~\cite{bijral}.

Let $p(x; \mu, \kappa)$ denote either a vMF density or a Watson density. We consider mixture models of $K$ different vMF densities or $K$ different Watson densities. Thus, a given unit norm observation vector $x$ has the \emph{mixture density}
\begin{equation}
  \label{eq:9}
  f\bigl(x; \set{\mu_j}_{j=1}^K, \set{\kappa_j}_{j=1}^K\bigr) := \nlsum_{j=1}^K\pi_jp(x; \mu_j, \kappa_j).
\end{equation}

Suppose we observe the set ${\cal X} = \{x_1,\dots, x_n \in \proj\}$ of i.i.d.\ samples drawn from~\eqref{eq:9}. Our aim is to infer the mixture parameters $(\pi_j,\mu_j,\kappa_j)_{j=1}^K$, where $\sum_j \pi_j=1$, $\pi_j \ge 0$, $\norm{\mu_j}=1$, and $\kappa_j \ge 0$ for an movMF and $\kappa_j \in \reals$ for a moW.

\subsection{EM algorithm}
A standard, practical approach to estimating the mixture parameters is via the Expectation Maximization (EM) algorithm~\cite{dlr77} applied to maximize the  
mixture log-likelihood for $\cal X$. Specifically, we seek to maximize
\begin{equation}
  \label{eq:17}
  \ell({\cal X}; \{\pi_j,\mu_j,\kappa_j\}_{j=1}^K) :=
  \nlsum_{i=1}^n \ln \Bigl(\nlsum_{j=1}^K \pi_j p(x; \mu_k, \kappa_j)\Bigr).
\end{equation}
To apply EM, first we use Jensen's inequality to compute the lower bound
\begin{equation}
  \label{eq:19}
  \ell({\cal X}; \{\pi_j,\mu_j,\kappa_j\}_{j=1}^K) \ge \nlsum_{ij} \beta_{ij} \ln \left(\pi_j p( x_i |  \m_j, \kappa_j) / \beta_{ij}\right).
\end{equation}
Then, the E-Step sets $\beta_{ij}$ to the \emph{posterior} probability (for $x_i$ given component $j$):
\begin{equation}
  \label{eq:18}
  \beta_{ij} := \frac{\pi_j p( x_i | \m_j,\kappa_j)}
  {\sum_l \pi_l p( x_i | \m_l, \kappa_l)}.
\end{equation}
With this choice of $\beta_{ij}$, the M-Step maximizes~\eqref{eq:19}, which is essentially just a maximum-likelihood problem, to obtain parameter updates. In particular, we obtain\\
\emph{M-Step for movMF:}
\begin{align}
  \label{eq:10}
  &\mu_j = \frac{r_j}{\norm{r_j}},\quad r_j = \nlsum_{i=1}^n\beta_{ij}x_i,\\
  \label{eq:11}
  &\kappa_j = A_p^{-1}(\rbar_j),\quad \rbar_j = \frac{\norm{r_j}}{\nlsum_{i=1}^n\beta_{ij}}.
\end{align}
\emph{M-Step for moW:}
\begin{align}
  \label{eq:12}
  &\mu_j = {s}_1^j\quad\text{if}\quad \kappa_j >
  0,\qquad\mu_j = {s}_p^j\quad\text{if}\quad\kappa_j < 0,\\
  \label{eq:13}
  &\kappa_j = g^{-1}(1/2,p/2,r_j),\quad\text{where}\quad
  r_j = \m_j^\top{S}^j\m_j
\end{align}
where ${s}_1^j$ denotes the top eigenvector corresponding to eigenvalue $\lambda_i(S_j)$  of the \emph{weighted-scatter matrix}
\begin{equation*}
  S^j = \frac{1}{\nlsum_{i=1}^n \beta_{ij}}\nlsum_{i=1}^n \beta_{ij}  x_i x_i^T.
\end{equation*}
For both movMF and moW, the component probabilities are as usual $\pi_j = \tfrac{1}{n}\nlsum_{i} \beta_{ij}$.
Iterating between~\eqref{eq:18} and the M-Steps we obtain an EM algorithm. Pseudo-code for such a procedure is shown below as Algorithm~\ref{algo:em}. 

\vskip8pt
\noindent\textbf{Hard Assignments.} To speed up EM, we can replace can E-Step~\eqref{eq:18} by the standard \emph{hard-assignment} rule:
\begin{equation}
  \label{eq:31}
  \beta_{ij} =
  \begin{cases}
    1, & \text{if}\ j = \argmax_{j'} \ln \pi_{l} + \ln p(x_i| \mu_{l},\kappa_{l}),\\
    0, & \text{otherwise}.
  \end{cases}
\end{equation}
The corresponding $M$-Step also simplifies considerably. Such hard-assignments maximize a lower-bound on the incomplete log-likelihood and yield \emph{partitional-clustering} algorithms. 

\vskip8pt
\noindent\textbf{Initialization.} For movMF, typically an initialization using spherical kmeans (spkmeans)~\cite{spkmeans} can be used. The next section presents arguments that explain why this initialization is natural for movMF. Similarly, for moW, an initialization based on diametrical kmeans~\cite{diametric} can be used, though sometimes even an spkmeans initialization suffices~\cite{sra.mow}.

\begin{algorithm}[t]\small
  \DontPrintSemicolon
  \caption{\small EM Algorithm for movMF and moW~\label{algo:em}}
  \KwIn{$x = \set{x_1,\ldots, x_n  : \text{ where each }  \norm{x_i}=1}$, $K$}
  \KwOut{Parameter estimates $\pi_j$, $\m_j$, and $\kappa_j$, for $1 \le j \le K$}
  Initialize $\pi_j, \m_j, \kappa_j$ for $1 \leq j \leq K$\;
  \While{not converged}{
    \emph{\{Perform the E-step of EM\}}\;
    \ForEach{$i$ and $j$}{
      Compute $\beta_{ij}$ using~(\ref{eq:18}) (or via~\eqref{eq:31})
    }
    \emph{\{Perform the M-Step of EM\}}\;
    \For{$j = 1$ to $K$}{
      $\pi_j \gets \frac{1}{n}\sum_{i=1}^n \beta_{ij}$\;
      For movMF: compute $\mu_j$ and $\kappa_j$ using~\eqref{eq:10} and \eqref{eq:11}\;
      For moW:  compute $\mu_j$ and $\kappa_j$ using~\eqref{eq:12} and \eqref{eq:13}\;
    }
  }
\end{algorithm}

\subsection{Limiting versions}
\label{sec:limit}
It is well-known that the famous k-means algorithm may be obtained as a limiting case of the EM algorithm applied to a mixture of Gaussians. Analogously, the spherical kmeans algorithm of~\cite{spkmeans} that clusters unit norm vectors and finds unit norm means (hence `spherical') can be viewed as the limiting case of a movMF. Indeed, assume that the priors of all mixture components are equal. Furthermore, assume that all the mixture components have equal concentration parameters $\kappa$ and let $\kappa \to \infty$. Under these assumptions, the E-Step~\eqref{eq:31} reduces to assigning a point $x_i$ to the cluster nearest to it, which here is given by the cluster with  whose centroid the given point has largest dot product. In other words, a point $x_i$ is assigned to cluster $k = \argmax_j x_i^T\mu_j$ because $\beta_{ik}\to 1$ and $\beta_{ij} \to 0$ for $j\neq k$ in~\eqref{eq:31}.

In a similar manner, the diametrical clustering algorithm of~\cite{diametric} also may be viewed as a limiting case of EM applied to a moW. Recall that the diametrical clustering algorithm groups together correlated and anti-correlated unit norm data vectors into the same cluster, i.e., it treats diametrically opposite points equivalently. Remarkably, it turns out that the diametrical clustering algorithm of~\cite{diametric} can be obtained as follows: Let $\kappa_j \to \infty$, so that for each $i$, the corresponding posterior probabilities~$\beta_{ij} \to \{0,1\}$; the particular $\beta_{ij}$ that tends to 1 is the one for which $(\m_j^\top x_i)^2$ is maximized in the E-Step; subsequently the M-Step \eqref{eq:12}, \eqref{eq:13} also simplifies, and yields the same updates as made by the diametrical clustering algorithm.

Alternatively, we can obtain diametrical clustering from the hard-assignment heuristic of EM applied to a moW where all mixture components have the same (positive) concentration parameter $\kappa$. Then, in the E-Step~(\ref{eq:31}), we can ignore $\kappa$ altogether, which reduces Alg.~\ref{algo:em} to the diametrical clustering procedure.

\subsection{Application: clustering using movMF}
Mixtures of vMFs have been successfully used in text clustering; see \cite{crchap} for a detailed overview. We recall below results of a two main experiments below: (i) simulated data; and (ii) Slashdot news articles. 

The key characteristic of text data is its high dimensionality. And for modeling clusters of such data using a movMF, the approximate computation of the concentration parameter $\kappa$ as discussed in Sec.~\ref{sec:kappa} is of great importance: without this approximation, the computation breaks down due to floating point difficulties.

For (i), we simulate a mixture of 4 vMFs in with $p=1000$ each, and draw a sample of 5000 data points. The clusters are chosen to be roughly of the same size and their relative mixing proportions are $(0.25,0.24,0.25,0.26)$, with concentration parameters (to one digiit) $(651.0,267.8,267.8,612.9)$, and random units vectors as means. This is the same data as the \texttt{big-mix} data in~\cite{crchap}. We generated the samples using the \texttt{vmfsamp} code (available from the author upon request). 

For (ii), we recall a part of the results of~\cite{crchap} on news articles from the Slashdot website. These articles are tagged and cleaned to retain 1000 articles that more clearly belong to a primary category / cluster. We report results on `Slash-7' and `Slash-6'; the first contains 6714 articles in 7 primary categories: business, education, entertainment, games, music, science, and internet; while the second contains 5182 articles in 6 categories: biotech, Microsoft, privacy, Google, security, and space. 

\paragraph{Performance evaluation.} There are several ways to evaluate performance of a clustering method. For the simulated data, we know the true parameters from which the dataset was simulated, hence we can compare the error in estimated parameter values. For the Slashdot data, we have knowledge of ``ground truth'' labels, so we can use the \emph{normalized mutual information (NMI)}~\cite{strehl02} (a measure that was also previously used to assess movMF based clustering~\cite{sra.vmf,crchap}) as an external measure of cluster quality. Suppose the predicted cluster labels are $\hat{Y}$ and the true labels are $Y$, then the NMI between $Y$ and $\hat{Y}$ is defined as
\begin{equation}
  \label{eq:5}
  \text{NMI}(Y, \hat{Y}) := \frac{I(Y, \hat{Y})}{\sqrt{H(Y)H(\hat{Y})}},
\end{equation}
where $I(\cdot,\cdot)$ denotes the usual mutual information and $H$ denotes the entropy~\cite{cover}. When the predicted labels agree perfectly with the true labels, then NMI equals 1; thus higher values of NMI are better.

\paragraph{Results on `bigsim'.} The results of the first experiment are drawn from~\cite{sra.vmf}, and are reported in Table~\ref{tab:bigsim}. From the results it is clear that on this particular simulated data, EM manages to recover the true parameters to quite a high degree of accuracy. Part of this reason is due to the high values of the concentration parameter: as $\kappa$ increases, the probability mass concentrates, which makes it easier to separate the clusters using EM. To compute the values in the table, we ran EM with soft-assignments and then after convergence used assignment (\ref{eq:31}).
\begin{table}[h]
  \centering
  \begin{tabular}{c|c|c}
    min $\mu^T\hat{\mu}$& max $\frac{|\kappa-\hat{\kappa}|}{\kappa}$ & max $\frac{|\pi-\hat{\pi}|}{\pi}$\\
    \hline
    0.994 & 0.006 & 0.002
  \end{tabular}
  \caption{Accuracy of parameter estimation via EM for movMF on the `bigsim' data. We report the worst values (the averages were better) seen across 20 different runs. For the estimated mean, we report the worst inner product with the true mean; for the concentration and mixture priors we report worst case relative errors.}
  \label{tab:bigsim}
\end{table}

\paragraph{Results on Slashdot.} These results are drawn from~\cite{crchap}. The results here are reported merely as an illustration, and we refer the reader to \cite{crchap} for more extensive results. We report performance of our implementation of Alg.~\ref{algo:em} (EM for movMF) against Latent Dirichlet Allocation (LDA)~\cite{blei2003latent} and a Exponential-family Dirichlet compound multinomial model (EDCM)~\cite{ecdm}. 

Table~\ref{tab:vmf} reports results of comparing Alg.~\ref{algo:em} specialized for movMFs against LDA and EDCM. As can be seen from the results, the vMF mixture leads to much higher quality clustering than the other two competing approaches. We did not test an optimized implementation (and used our own \textsc{Matlab} implementation), but note anecdotally that the EM procedure was 3--5 times faster than the others. 
\begin{table}[h]
  \centering
  \begin{tabular}{c|ccc}
    Dataset & moVMF & LDA & ECDM\\
    \hline
    Slash-7 & 0.39 & 0.22 & 0.31\\
    Slash-6 & 0.65 & 0.36 & 0.46
  \end{tabular}
  \caption{Comparison of NMI values of moVMF versus LDA and ECDM (derived from~\cite{crchap}).}
  \label{tab:vmf}
\end{table}

\subsection{Application: clustering using moW}
\begin{figure}
  \centering
\hspace*{-20pt}\includegraphics[scale=0.3]{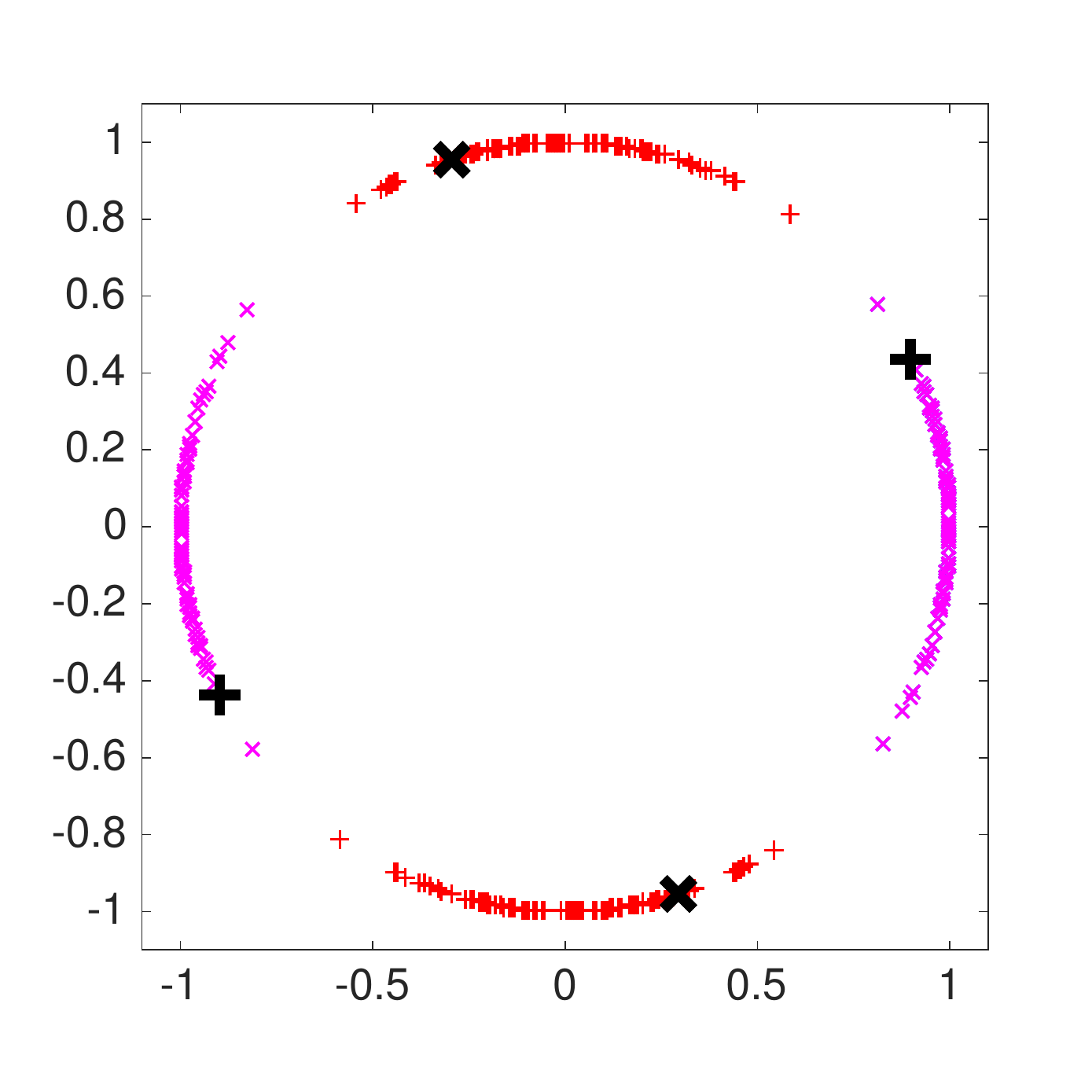}~\includegraphics[scale=0.3]{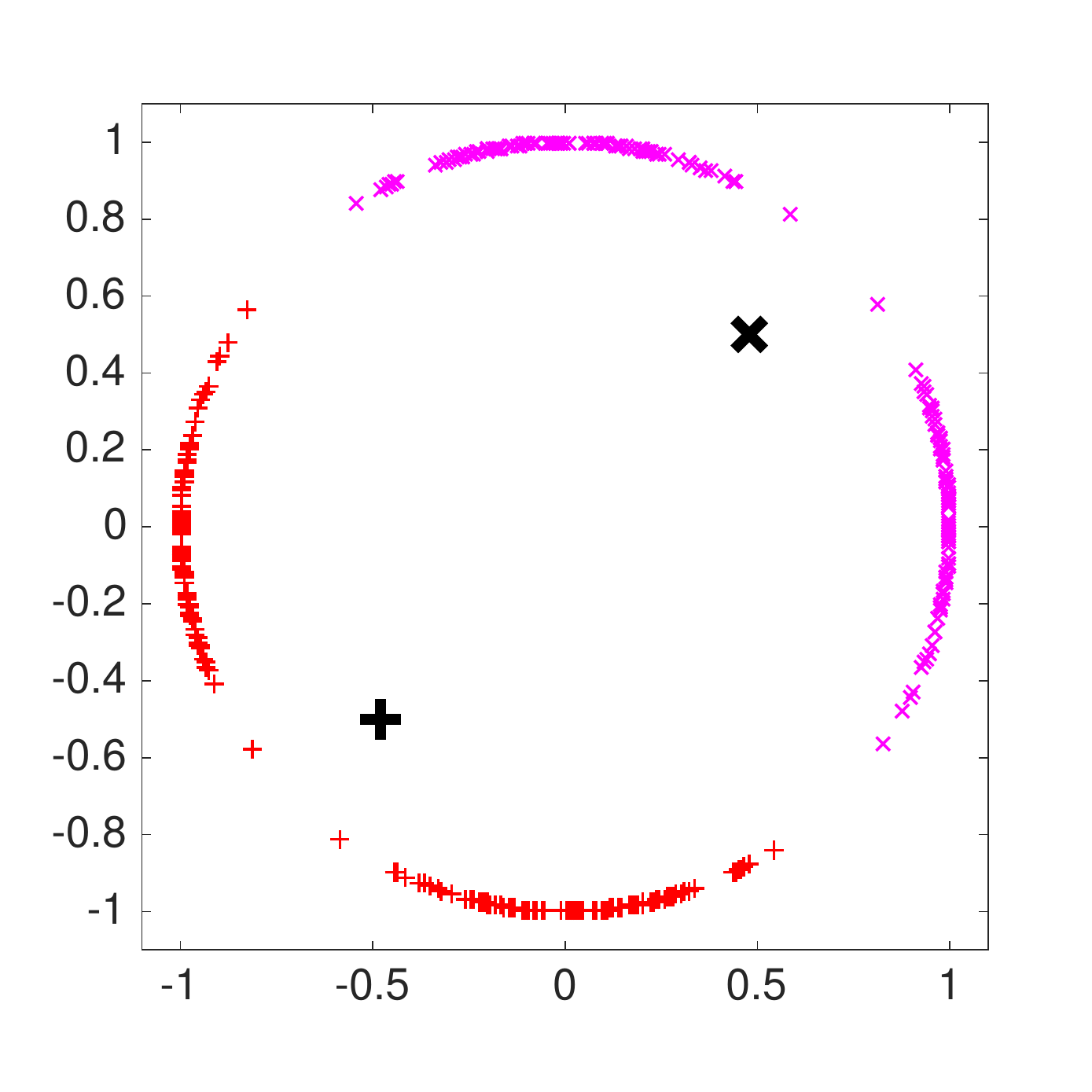}~\includegraphics[scale=0.3]{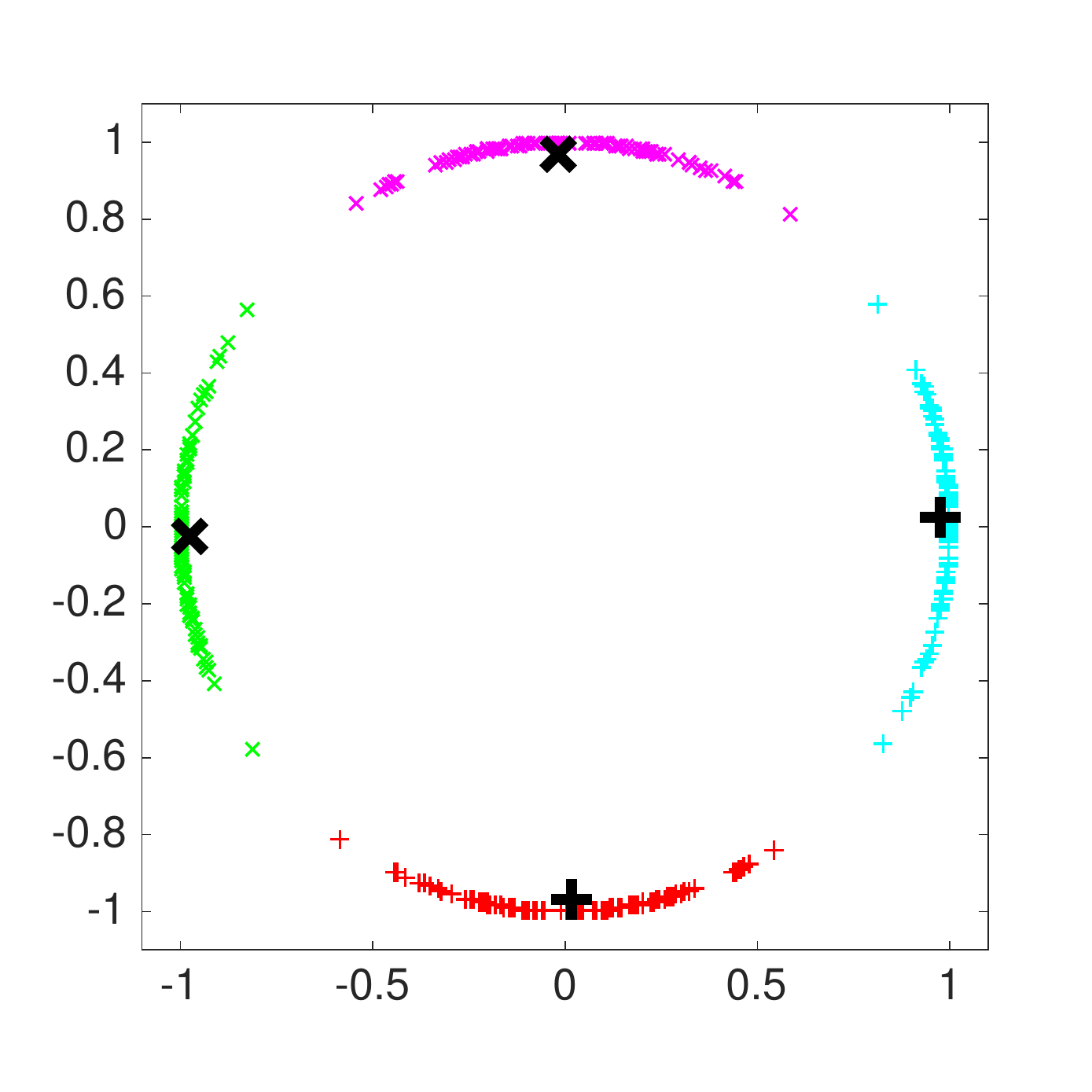}
  \caption{The left panel shows axially symmetric data that has two clusters (centroids are indicated by '+' and  'x'). The middle and right panel shows clustering yielded by (Euclidean) K-means (note that the centroids fail to lie on the circle in this case) with $K=2$ and $K=4$, respectively. Diametrical clustering recovers the true clusters in the left panel.}
  \label{fig:diam}
\end{figure}

Figure~\ref{fig:diam} shows a toy example of axial data. Here, the original data has two clusters (leftmost panel of Fig.~\ref{fig:diam}). If we cluster this data into two clusters using Euclidean kmeans, we obtain the plot in the middle panel; clustering into 4 groups using Euclidean kmeans yields the rightmost panel. As is clear from the figure, Euclidean kmeans cannot discover the desired structure, if the true clusters are on axial data. The diametrical clustering algorithm of \cite{diametric} discovers the two clusters (leftmost panel), which also shows the mean vectors $\pm\mu$ for each cluster. Recall that as mentioned above, the diametrical clustering method is obtained as the limiting case of EM on moW.

\section{Conclusion}
We summarized a few distributions from directional statistics that are useful for modeling normalized data. We focused in particular on the von Mises-Fisher distribution (the ``Gaussian'' of the hypersphere) and the Watson distribution (for axially symmetric data). For both of these distributions, we recapped maximum likelihood parameter estimation as well as mixture modeling using the EM algorithm. For extensive numerical results on clustering using mixtures of vMFs, we refer the reader to the original paper~\cite{sra.vmf}; similarly, for mixtures of Watsons please see~\cite{sra.mow}. The latter paper also describes asymptotic estimates of the concentration parameter in  detail. 

Now directional distributions are widely used in machine learning (Sec.~\ref{sec:rel} provides some pointers to related work), and we hope the brief summary provided in this chapter helps promote wider understanding about these. In particular, we hope to see more exploration of directional models in the following important subareas: Bayesian models, Hidden Markov Models using directional models, and deep generative models.


\bibliographystyle{abbrv}

\end{document}